\documentclass[pdflatex,sn-nature]{sn-jnl}

\makeatletter
\@ifundefined{SN@ands}{}{\renewcommand{\SN@ands}{ and }}
\makeatother

\usepackage{graphicx}%
\usepackage{multirow}%
\usepackage{amsmath,amssymb,amsfonts}%
\usepackage{amsthm}%
\usepackage{mathrsfs}%
\usepackage[title]{appendix}%
\usepackage{xcolor}%
\usepackage{textcomp}%
\usepackage{manyfoot}%
\usepackage{booktabs}%
\usepackage{algorithm}%
\usepackage{algorithmicx}%
\usepackage{algpseudocode}%
\usepackage{listings}%
\usepackage{bbm}
\usepackage{threeparttable}
\usepackage{dsfont}
\theoremstyle{thmstyleone}%
\theoremstyle{thmstyletwo}%

\theoremstyle{thmstylethree}%

\begin{document}

\title[Article Title]{
\centering
Enhancing Time-Series Anomaly Detection \\
by Integrating Spectral-Residual \\
Bottom-Up Attention with Reservoir Computing}

\author[1]{\fnm{Hayato} \sur{Nihei}}\email{\nolinkurl{s2581033DL@s.chibakoudai.jp (HN)}}

\author[1,2,3,4]{\fnm{Sou} \sur{Nobukawa}}\email{nobukawa@it-chiba.jp (SN)}

\author[4,5]{\fnm{Yusuke} \sur{Sakemi}}

\author[4,5,6]{\fnm{Kazuyuki} \sur{Aihara}}

\affil[1]{\orgdiv{Graduate School of Information and Computer Science}, \orgname{Chiba Institute of Technology}, \orgaddress{\street{2-17-1 Tsudanuma}, \city{Narashino}, \postcode{275-0016}, \state{Chiba}, \country{Japan}}}

\affil[2]{\orgdiv{Department of Computer Science}, \orgname{Chiba Institute of Technology}, \orgaddress{\street{2-17-1 Tsudanuma}, \city{Narashino}, \postcode{275-0016}, \state{Chiba}, \country{Japan}}}

\affil[3]{\orgdiv{Department of Preventive Intervention for Psychiatric Disorders}, \orgname{National Center of Neurology and Psychiatry}, \orgaddress{\street{4-1-1 Ogawa-Higashi}, \city{Kodaira}, \postcode{187-8551}, \state{Tokyo}, \country{Japan}}}

\affil[4]{\orgdiv{Research Center for Mathematical Engineering}, \orgname{Chiba Institute of Technology}, \orgaddress{\street{2-17-1 Tsudanuma}, \city{Narashino}, \postcode{275-0016}, \state{Chiba}, \country{Japan}}}

\affil[5]{\orgdiv{International Research Center for Neurointelligence}, \orgname{The University of Tokyo}, \orgaddress{\street{7-3-1 Hongou}, \city{Bunkyo ku}, \postcode{113-8654}, \state{Tokyo}, \country{Japan}}}

\affil[6]{\orgdiv{Institute for AI and Beyond}, \orgname{The University of Tokyo}, \orgaddress{\street{7-3-1 Hongou}, \city{Bunkyo ku}, \postcode{113-8654}, \state{Tokyo}, \country{Japan}}}

\abstract{
Reservoir computing (RC) establishes the basis for the processing of time-series data by exploiting the high-dimensional spatiotemporal response of a recurrent neural network to an input signal.
In particular, RC trains only the output layer weights.
This simplicity has drawn attention especially in Edge Artificial Intelligence (AI)  applications.
Edge AI enables time-series anomaly detection in real time, which is important because detection delays can lead to serious incidents.
However, achieving adequate anomaly-detection performance with RC alone may require an unacceptably large reservoir on resource-constrained edge devices.
Without enlarging the reservoir, attention mechanisms can improve accuracy, although they may require substantial computation and undermine the learning efficiency of RC.
In this study, to improve the anomaly detection performance of RC without sacrificing learning efficiency, we propose a spectral residual RC (SR-RC) that integrates the spectral residual (SR) method—a learning-free, bottom-up attention mechanism—with RC.
We demonstrated that SR-RC outperformed conventional RC and logistic-regression models based on values extracted by the SR method across benchmark tasks and real-world time-series datasets.
Moreover, because the SR method, similarly to RC, is well suited for hardware implementation, SR-RC suggests a practical direction for deploying RC as Edge AI for time-series anomaly detection.
}


\maketitle

\section*{Introduction}

%First paragraph============
Significantly large volumes of time-series data are continuously obtained from diverse real-world sources, including sensor networks and industrial systems \cite{atzori2010internet,chong2003sensor,madakam2015internet,8558534}.
Effective leveraging of these data is critically important across a wide range of fields, including medicine \cite{fried2017online}, finance \cite{sezer2020financial,tsay2005analysis}, and climate science \cite{mudelsee2019trend}.
However, current practice typically relies on centralised server-side processing; as data volumes increase and the available bandwidth is constrained, transferring data to remote servers introduces delays that are difficult to avoid \cite{su2022ai,chen2015efficient,shi2016edge,wang2022big}.
These delays hinder the early detection of serious risks, such as system failures, fraudulent activities, and health problems, and may lead to major incidents \cite{chandola2009anomaly,schmidl2022anomaly,Ahmed2016finance,ahmed2016IT,fernando2021deep}.
In particular, because many risks manifest as anomalies in time series data, data transfer delays pose a major challenge for time series anomaly detection \cite{cook2019anomaly,blazquez2021review,fahim2019anomaly}.
To address this challenge, detection and analysis is performed near the data acquisition point, thereby minimising the data transfer, which is the principal source of delay \cite{yang2024edgebench,khan2019edge,kong2022edge}.
Edge artificial intelligence (AI), which executes computations on edge devices, supports this approach by reducing data transfer \cite{gill2025edge,SINGH202371,chang2021survey,tekin2024review,deng2020edge}.
By processing data on devices, edge AI achieves low latency while maintaining high anomaly detection performance and is therefore regarded as an important technology for time-series anomaly detection \cite{demedeiros2023survey,trilles2024anomaly,debauche2020new}.
Within edge AI, reservoir computing (RC), which exhibits a high predictive performance and learning efficiency, has attracted attention in recent years, and its application to time-series anomaly detection has been widely studied \cite{yan2024emerging,zhang2023survey}.

%Second paragraph============
RC, a derivative of a recurrent neural network (RNN), offers high predictive accuracy and learning efficiency, and achieves notable results across diverse domains of time-series anomaly detection \cite{lukovsevivcius2009reservoir,cucchi2022hands,tanaka2019recent,nakajima2020physical}.
Similar to a conventional RNN, RC consists of an input layer, a recurrent layer, and an output layer, with the recurrent layer referred to as the reservoir layer in RC \cite{jaeger2001echo}.
In RC, the input signal is mapped to a high-dimensional dynamical system within the reservoir, where the spatiotemporal response functions as a kernel that extracts features from the input signal.
Consequently, even with randomly fixed internal connections in the recurrent layer, learning can be achieved by simply linearly combining the spatiotemporal responses in the output layer \cite{cucchi2022hands}.
Representative anomaly detection techniques using RC include RC-based autoencoders for feature extraction and classification, reconstruction error approaches to model normal data, and techniques for identifying anomalies through prediction errors\cite{Chen2018ImbalancedDE,kato2024reconstructive,de2022novel}.
These methods have achieved consistent performances across a range of tasks, such as detecting abnormal heartbeats\cite{Chen2018ImbalancedDE}, fault detection in sensor data \cite{long2019novel}, and anomaly detection in embedded devices \cite{carta2022efficient,chang2009mote}.
However, for these RC-only methods, achieving adequate performance may require an unacceptably large reservoir \cite{lukovsevivcius2009reservoir,lukovsevivcius2012practical}.
Therefore, when available physical resources are limited, as in implementations on edge devices, the reservoir size is often constrained, and it remains unclear whether such RC-based anomaly detection methods can achieve their intended performance under these constraints \cite{tanaka2019recent,nakajima2020physical,stepney2024physical}.

%Third paragraph=================
Improving the training strategy is one way to enhance RC-based anomaly-detection performance without increasing reservoir size.
There are two primary learning methods for anomaly detection \cite{nassif2021machine}: one that exclusively utilises normal data and one that utilises both normal and anomalous data \cite{zamanzadeh2024deep,gornitz2013toward}.
Many proposed anomaly detection methodologies exclusively employ normal data for training because of the scarcity of anomalous data \cite{Hassan2021ARO,ruff2018deep}.
Models trained on sufficient normal data can identify anomalies based on indicators such as reconstruction error or prediction errors \cite{pimentel2014review,munir2018deepant,yin2020anomaly}.
When the input signals contain anomalous data, the model's predictions or reconstructions exhibit significant errors compared to normal data, enabling anomaly detection through threshold settings or statistical methods.
However, previous research has demonstrated that including anomalous data in the training process is beneficial for enhancing model performance \cite{ruff2020rethinking,pang2019deep,liznerski2020explainable}.
Incorporating anomalous data allows the model to learn the boundary between normal and abnormal data more distinctly and capture the specific characteristics of anomalies. 
This approach improves the generalisation and robustness of the model, leading to highly accurate and reliable anomaly detection.

%Fourth paragraph=================
In addition to the learning method, integrating an attention mechanism with RC is another promising approach for enhancing anomaly detection performance.
The attention mechanism is a breakthrough technique proposed in deep learning and has improved anomaly detection performance in many domains \cite{ma2024research,choi2021deep,niu2021review}.
These methods compute an attention distribution from the input and hidden states of the network and apply it to the input to emphasise important information, such as anomalies, while suppressing less relevant data \cite{ma2024research,choi2021deep,brauwers2021general,de2022attention}.
Although distinct from anomaly detection, research in the field of time-series forecasting has demonstrated an enhanced learning performance through the combination of RC with these mechanisms \cite{sakemi2024learning,koster2023attention,konishi2023insar}.
These findings suggest that incorporating an attention mechanism into RC could enhance the anomaly detection performance.
However, these methods often result in increased training times \cite{zhuang2023survey,de2022attention}.
In particular, training that computes gradients via backpropagation \cite{rumelhart1986learning} or backpropagation through time (BPTT) \cite{werbos2002backpropagation} and updates parameters via gradient descent incurs high computational load and energy consumption \cite{williams1990efficient,schmidhuber2015deep,sze2017efficient}.

%Five paragraph=================
The attention mechanism described above is classified in neuroscience as top-down attention and differs substantially from bottom-up attention with respect to the requirements for learning \cite{chica2013two,meyer2018exogenous,carrasco2011visual,katsuki2014bottom}. 
Top-down attention endogenously and voluntarily guides attention to specific locations or objects based on goals and prior knowledge \cite{chica2013two,thiele2018neuromodulation}.
The attention mechanisms previously used in RC mimic top-down attention because they dynamically allocate attention to highly relevant information in the input according to the task \cite{niu2021review}.
In contrast, bottom-up attention guides focus on  stimuli that stand out from their surroundings among multiple observed stimuli \cite{chica2013two,duncan1989visual,carrasco2011visual,katsuki2014bottom,nothdurft1993role,treisman1980feature}.
These stimuli, characterised by features such as colour, edges, and frequency components, do not require prior knowledge or pretraining for fixation \cite{knierim1992neuronal,schein1990spectral,tanaka1986analysis}. 
This process prioritises salient information that is likely to aid the task while filtering out irrelevant noise \cite{pratte2013attention,sperling1960information,Cowan2023TheRB}.
A saliency map is a theoretical computational mechanism that models bottom-up attention, namely, the stimulus-driven, involuntary orienting of attention \cite{itti2000saliency,itti1998model,hou2007saliency}.
This mechanism is widely used in computer vision to compare the information at each pixel in an input image and to treat regions with larger deviations as more salient \cite{borji2019salient}.
In addition, saliency maps can detect saliency not only from images, but also from time series data \cite{ren2019time,Pan2020Series}.
Among these methods, the spectral residual (SR) method is a particularly efficient algorithm for generating saliency maps from the frequency components of time series data \cite{hou2007saliency,ren2019time}. 
When combined with convolutional neural networks (CNNs), the SR method exhibits high performance in time-series anomaly detection, indicating that saliency maps work well with these neural network architectures \cite{ren2019time}.

%Sixth paragraph=================
The RC and SR methods are highly amenable to hardware implementations.
The high performance of RC arises from the fact that in the reservoir layer, input signals are transformed into nonlinear, high-dimensional representations, and useful features can be extracted.
Accordingly, the transformation performed by the reservoir layer can be realised by using other nonlinear dynamical systems \cite{tanaka2019recent,nakajima2020physical,stepney2024physical}.
In practice, RC has been implemented in spintronic oscillators and silicon photonic integrated circuits, and high energy efficiency has been reported \cite{tanaka2019recent,nakajima2020physical,marrows2024neuromorphic,van2017advances}.
Moreover, because the SR method is based on bottom-up attention and has no parameters that require training such as  backpropagation or BPTT.
Its primary computations are the fast Fourier transform (FFT) and its inverse (IFFT), which are widely implemented in digital signal processors (DSPs) and application-specific integrated circuits (ASICs) and offer high energy efficiency \cite{garrido2022survey,garrido2020evolution,noor2018design}.

%Seventh paragraph=================
As noted previously, RC—a leading candidate for edge AI—has attracted particular attention in time-series anomaly detection. 
However, when physical resources are constrained—as in on-edge devices—current RC-based anomaly detection methods raise concerns.
Moreover, employing top-down attention mechanisms can undermine RC’s advantage in terms of learning efficiency.
Therefore, we hypothesise that integrating a bottom-up attention mechanism that requires no training into RC can improve anomaly detection performance without degrading learning efficiency.
To verify this hypothesis, we propose a spectral residual RC (SR-RC) architecture, incorporating a bottom-up attention mechanism into RC, rather than a learning based SR method. 
Specifically, the SR-RC model includes two variants: one that inputs only the saliency map into RC and one that inputs both the saliency map and original time series data.
Furthermore, training with time series that include anomalies enables the SR-RC to capture anomalous characteristics and distinguish them clearly from normal data.
In addition, the SR-based bottom-up attention mechanism has no parameters that require training and is expected to operate readily on existing signal-processing infrastructures.
Accordingly, similar to the conventional RC, the SR-RC architecture is expected to be highly amenable to hardware implementation.
Consequently, the SR-RC architecture, which uses a much smaller reservoir, achieves performance comparable to RC alone and offers a new option for employing RC as edge AI in time-series anomaly detection.
In the following sections, we focus on the architectural characteristics and anomaly detection performance of the SR-RC architecture and evaluate the performance of anomaly-detection benchmark tasks \cite{Lai2021RevisitingTS} and the Yahoo! Webscope S5 dataset \cite{laptev2015yahoo}.

%%------------------------------------------------
% results
%%------------------------------------------------
\section*{Results}
\begin{figure}[t]
\centering
\includegraphics[width=\linewidth]{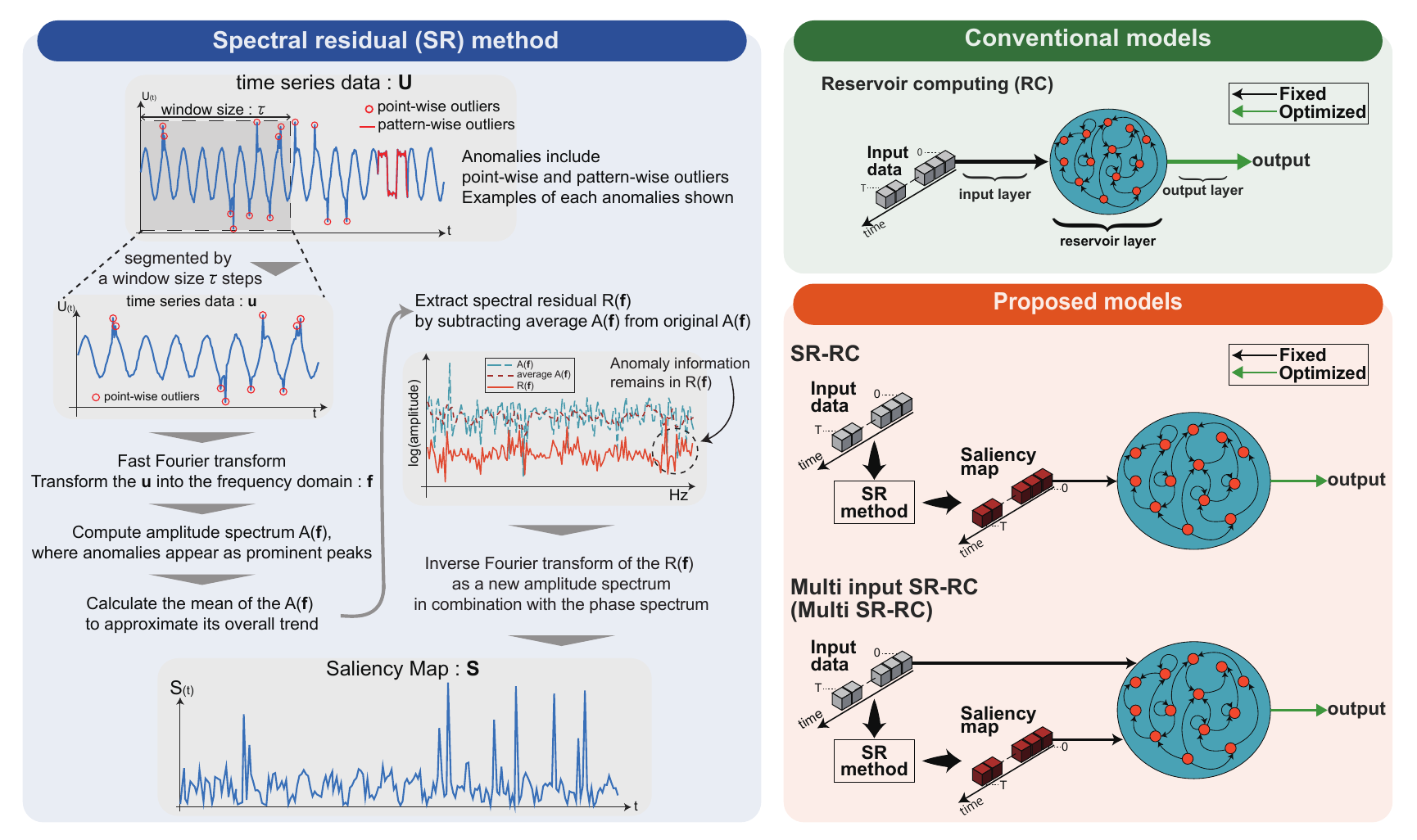}
\caption{Structure of the Spectral Residual (SR) method and the proposed models. 
The left side of the figure depicts the processing procedure of the SR method. 
The SR method applies a Fourier transform to the time series data, computes the spectral residual, and generates a saliency map through an inverse Fourier transform (see Methods for details).
As illustrated in the upper right of the figure, the reservoir computing (RC) system comprises an input layer, a reservoir layer, and an output layer. 
The conventional model inputs only the time series data into the RC. 
In contrast, the lower-right panel presents the proposed models: one that inputs only the saliency map and a multi-input model that uses both the original time series data and saliency map (see Methods for details).
}
\label{ConceptualFigure_ProposedModel}
\end{figure}

As shown in Fig. \ref{ConceptualFigure_ProposedModel}, the SR-RC architecture proposed in this study comprises two main components: the application of the SR method to time-series data for saliency-map generation and anomaly detection using RC.
The SR-RC architecture is designed to utilise either the saliency map alone or both the saliency map and raw time-series data without saliency map processing (referred to as the original time-series data in this paper).
Specifically, we propose two models: SR-RC, which employs a reservoir that receives only the saliency map, and multi-input SR-RC (Multi-SR-RC), which employs a reservoir that receives both the saliency map and original time-series data.

In the SR-RC, the reservoir is driven solely by the saliency map obtained from the original time-series data.
Using the saliency map as input allowed RC to detect anomalies more readily than using the original time series data.
In this study, we considered one-dimensional time series data; accordingly, the saliency map was one-dimensional.
The internal state of the SR-RC at time step $t\,(0 \le t \le T)$ is defined as:
\begin{align}
    \mathbf{x}_{t} = (1-\alpha)\,\mathbf{x}_{t-1} + \alpha \cdot \tanh\left(\mathbf{W}_{S}S_{t} + \mathbf{W}\,\mathbf{x}_{t-1}\right). \label{eq:SR-RC}
\end{align}
Here, $\mathbf{x}_{t} \in \mathbb{R}^{N}$ represents the internal state of the RC, consisting of $N$ neurones; $S_{t} \in \mathbb{R}$ is the one-dimensional saliency map obtained by applying the SR method to the original time-series data.
To compute the saliency map, we slide a window of size $\tau \in \mathbb{N}$ along the original time series, apply the SR method to each window, and concatenate the outputs to obtain $\mathbf{S}=(S_{1},S_{2},\ldots,S_{T})\in \mathbb{R}^{T}$, whose length equals $T$, the length of the original time-series data (see Methods).
The parameter $\alpha$ $(0 \le \alpha \le 1)$ is the leak rate, $\mathbf{W}_{S} \in \mathbb{R}^{N}$ denotes the input-coupling weights for $S_{t}$, and $\mathbf{W} \in \mathbb{R}^{N \times N}$ denotes the recurrent-coupling weights.
Both $\mathbf{W}$ and $\mathbf{W}_{S}$ are fixed.

The SR method was designed to detect change points caused by anomalies. 
However, when an anomaly persists, the saliency map highlights only the beginning and end of the anomaly, making it difficult for the SR-RC to capture the entire anomaly period.
To address this issue, Multi-SR-RC takes both the original time-series data and the saliency map as inputs.
This configuration allowed the model to extract information regarding the onset of anomalies from the saliency map and information about the duration of anomalies from the original time-series data.
In Multi SR-RC, the internal state of the reservoir layer at time step \(t\) is defined as
\begin{align}
    \mathbf{x}_{t} = (1-\alpha)\,\mathbf{x}_{t-1} + \alpha \cdot \tanh\left(\mathbf{W}_{\mathrm{in}}\,u_{t} + \mathbf{W}_{\mathrm{S}}\,S_{t} + \mathbf{W}\,\mathbf{x}_{t-1}\right). \label{eq:Multi-SR-RC}
\end{align}
Here, $u_t$ denotes the one-dimensional original time-series data, and $S_t$ is its saliency map. 
The input-coupling weights $\mathbf{W}_{\mathrm{in}}\in \mathbb{R}^{N}$ for $u_t$, $\mathbf{W}_{S}$ for $S_{t}$, and the recurrent-coupling weight matrix $\mathbf{W}$ are fixed.

\subsection*{Synthetic time-series anomaly-detection benchmark task}
\begin{figure}[t]
\centering
\includegraphics[width=\linewidth]{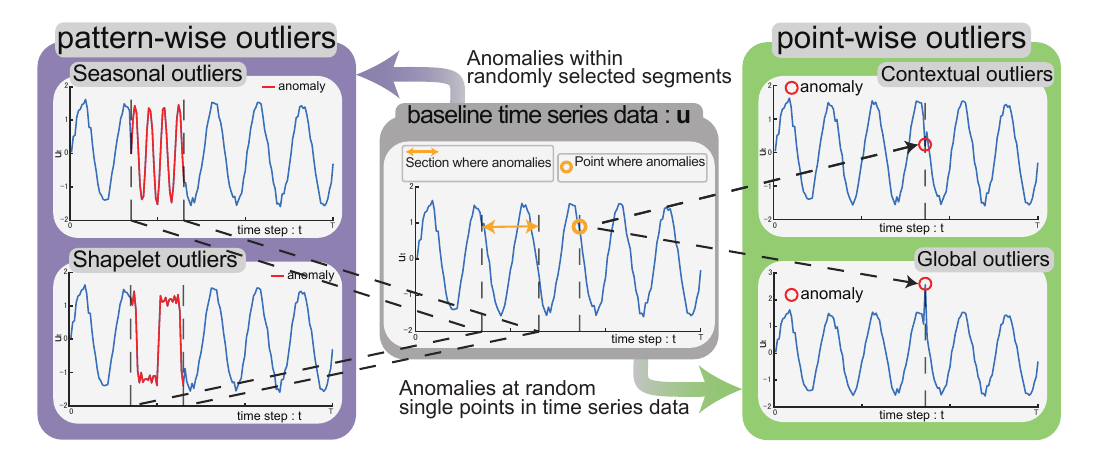}
\caption{Baseline time series data and introduced anomalies. 
 The central part of the figure shows the baseline time series data. 
 Anomalies are introduced at randomly selected points or intervals within this central time series. 
 The bottom-right of the figure illustrates global outliers, which are anomalies introduced by inserting spikes at randomly selected points in the time series. 
 The top-right of the figure depicts contextual outliers, where randomly selected points in the time series deviate from their surrounding values. 
 The bottom-left of the figure shows shapelet outliers, which are introduced by selecting random intervals in the time series and replaced by different waveforms. 
 The top-left of the figure illustrates seasonal outliers, which are created by selecting random intervals in the time series and altering the frequency of the underlying wave.}
\label{ConceptualFigure_BenchMarkTask}
\end{figure}

To evaluate the anomaly detection performance of the proposed SR RC architecture, we employed a benchmark task for time series anomaly detection.
This benchmark, originally proposed by Lai et al.\cite{Lai2021RevisitingTS}, injects synthetic anomalies into anomaly free baseline time series data, as illustrated in Fig. \ref{ConceptualFigure_BenchMarkTask}.
The injected anomalies can be classified into two categories: point-wise outliers and pattern-wise outliers.
Point-wise outliers were instantaneous anomalies.
In this study, we used two types of point-wise outliers, as shown in the bottom-right and top-right panels of Fig. \ref{ConceptualFigure_BenchMarkTask}: global outliers and contextual outliers, respectively (see Methods for details).
This task requires the detector to capture both the exact time points of an anomaly's occurrence and the surrounding normal behavior and to evaluate the disparity between them.
Pattern-wise outliers persist over a contiguous interval.
In this study, we use two types of Pattern-wise outliers, as depicted in the bottom-left and top-left panels of Fig. \ref{ConceptualFigure_BenchMarkTask}: shapelet outliers and seasonal outliers, respectively (see Methods for details).
The onset of an anomaly, its behavior throughout the interval, and its termination must be identified in this task, thereby capturing the entire anomalous segment.
The anomalies to be injected are determined by a Poisson process in which the probability that an anomalous value occurs at each time step is $\delta$ ($0<\delta \le 1$).
The baseline time series used in the benchmark consisted of two cases: one composed of a single sine wave and the other composed of a superposition of four sine waves (see Methods).

\begin{figure}[t]
\centering
\includegraphics[width=\linewidth]{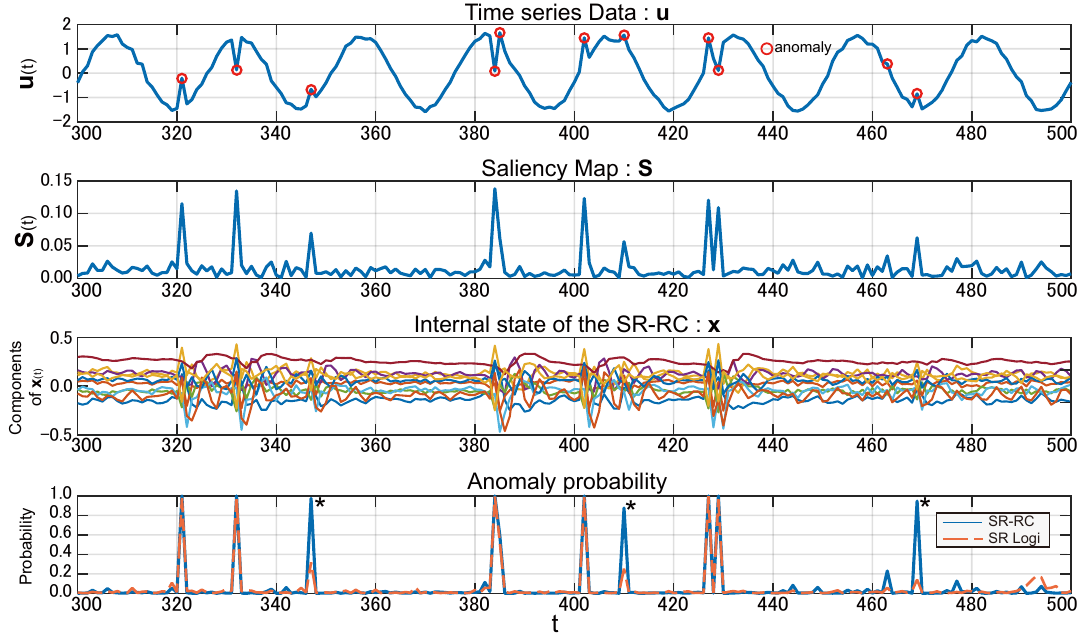}
\caption{Simulation results of SR-RC for point-wise outliers.
 From top to bottom, the figure displays the one-dimensional time-series data with contextual outliers, the corresponding saliency map computed from this data, the dynamics of 10 neurones in the reservoir layer of the SR-RC model, and the anomaly probabilities predicted by both the SR-RC and SR-Logi models.
 In this experiment, the anomaly occurrence probability ($\delta$) for contextual outliers was set to 0.05, and the SR-RC model comprised 100 neurones. 
 The reservoir layer exhibited transient fluctuations when spikes were input, depending on the magnitude of the spikes in the saliency map.
 In the bottom panel, asterisks denote cases in which SR-RC predicts a high anomaly probability for subtle spikes in the saliency map, whereas the logistic-regression model does not.
}
\label{internalState}
\end{figure}

Logistic regression is one option for edge-AI; however, the time-history effect of RC is useful for anomaly detection and is expected to outperform logistic regression.
To verify this, we first compared SR-RC with SR-Logi, a logistic regression model that uses only saliency map $\mathbf{S}$ as its feature vector (see Methods for details).
Figure~\ref{internalState} shows, from top to bottom, the time-series data containing contextual outliers, the saliency map $\mathbf{S}$ computed from these data using the SR method, the internal reservoir dynamics of SR-RC when $\mathbf{S}$ is used as the input, and the anomaly probability outputs from SR-RC and SR-Logi.
In the bottom panel—around $t{=}350,410,470$—SR-RC identifies weak spikes in the saliency map as anomalies, whereas the logistic regression model does not.
These observations indicate that the time-history effect of the RC improves the anomaly detection performance and that combining the SR method with RC is appropriate.

\begin{figure}[t]
    \centering
    \includegraphics[width=\linewidth]{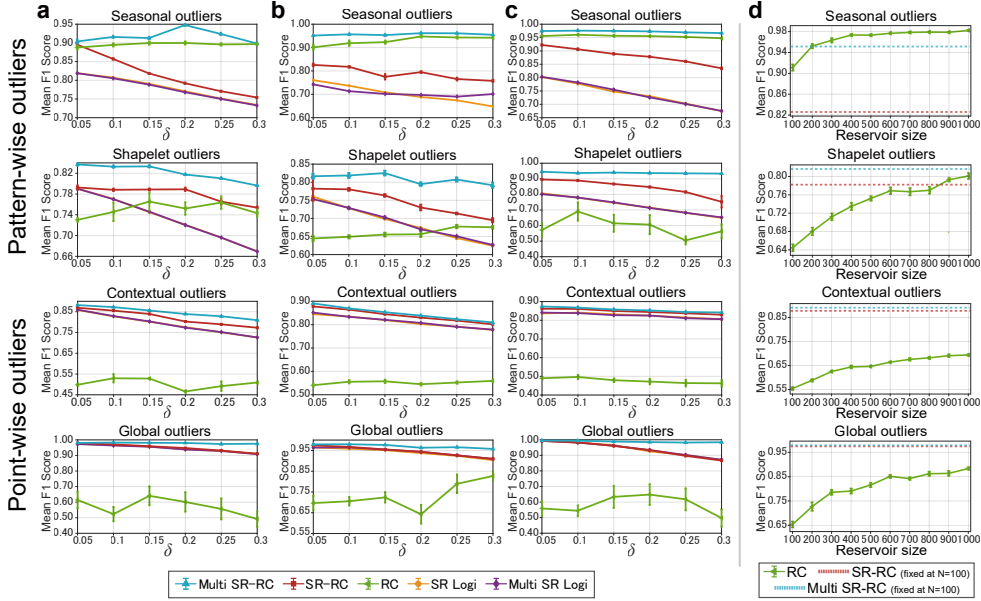}
        \caption{$\mathbf{a}$ Performance comparison on the benchmark task when the baseline time series is a single sine wave.
        $\mathbf{b}$ Performance comparison when the baseline is the sum of four sine waves.
        $\mathbf{c}$ Performance comparison when the baseline is the sum of four sine waves with irrational frequency ratios.
        For $\mathbf{a}$–$\mathbf{c}$, the x-axis shows the anomaly probability at each time step; the mean F1 score is averaged over 10 runs, and error bars denote the standard error.
        $\mathbf{d}$ For the four-sine baseline, the mean F1 score obtained with RC as a function of reservoir size. 
        The mean is averaged over 10 runs, and error bars denote the standard error. 
        Horizontal dashed lines (light blue and red, respectively) indicate the mean F1 scores of Multi-SR-RC and SR-RC at a fixed reservoir size of 100.
        In $\mathbf{a}$–$\mathbf{d}$, from top to bottom, the baseline time series is injected with seasonal, shapelet, contextual, and global outliers.
                }
    \label{AllResults}
\end{figure}

Next, we compared the two proposed models (Multi-SR-RC and SR-RC) with conventional logistic-regression models (SR-Logi and Multi-SR-Logi) and a conventional RC model across benchmark tasks.
Here, Multi-SR-Logi employs both the saliency map and the original time-series data as features.
Figure~\ref{AllResults}a–c shows the mean F1-scores (see Methods for definition) obtained when the benchmark’s baseline time series were set as a single sinusoid (Fig.\ref{AllResults}a), the sum of four sinusoids (Fig.\ref{AllResults}b), and a quasi-periodic series formed by summing four sinusoids with irrational pairwise frequency ratios (Fig.\ref{AllResults}c) (see Methods for details of the experimental setup).
Across all tasks, the reservoir size for the proposed models and the conventional RC model was fixed at $N=100$.
Across these comparisons, Multi-SR-Logi serves as the logistic-regression model that uses both the original time-series data and the saliency map as features. 
For each benchmark task, we vary the per-time-step anomaly probability as $\delta = {0.05, 0.10, \ldots, 0.30}$.
The results are shown in Fig.\ref{AllResults}a–c, Multi-SR-RC consistently achieved the highest performance across anomaly types.
SR-RC was the next best method after Multi-SR-RC in all tasks except seasonal outliers where RC outperformed SR-RC.

To assess whether an appropriate architectural design can improve accuracy while reducing the dependence on reservoir size, we compared the conventional RC at varying reservoir sizes against the two proposed models at fixed $N=100$ on the same benchmarks. 
Figure~\ref{AllResults} (d) summarises the results.
For point-wise outliers, both proposed models outperformed the conventional RC even when RC used a reservoir that was 10 times larger.
For shapelet outliers, Multi-SR-RC showed the same trend as for point-wise outliers, and SR-RC also surpassed conventional RC even when RC used a reservoir eight times larger.
For seasonal outliers, RC required a reservoir about two times larger to achieve a performance comparable to that of Multi-SR-RC.
Together, these results indicate that an SR-based bottom-up attention mechanism with an appropriate architectural design improves anomaly-detection accuracy and achieves performance comparable to RC alone even with a small reservoir.

\subsection*{Real-world time-series anomaly-detection task}

\begin{figure}[t]
    \centering
    \includegraphics[width=\linewidth]{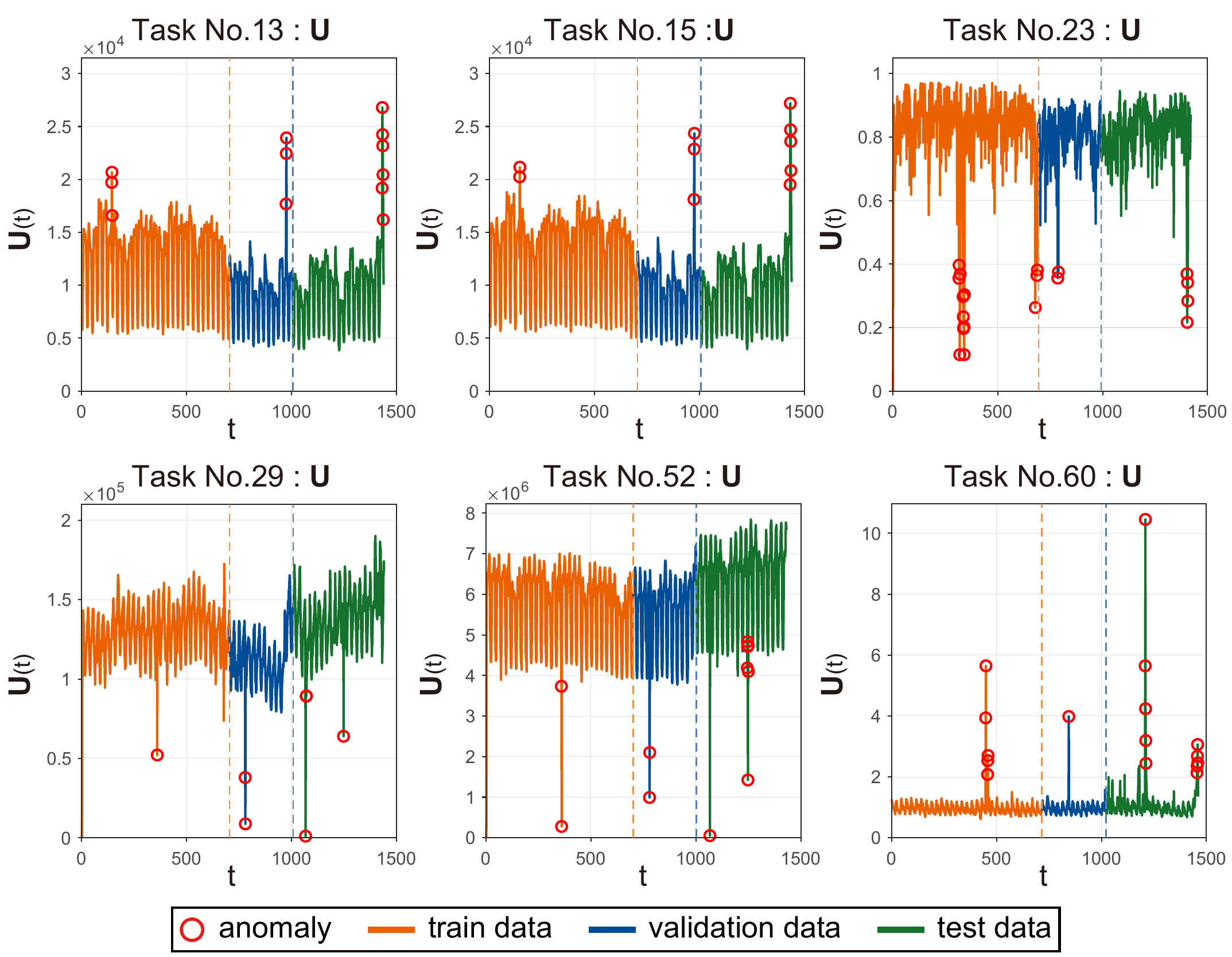}
        \caption{Overview of the Yahoo! Webscope S5 dataset used for real-world time-series anomaly-detection tasks.
        The dataset is divided into four benchmarks (A1–A4). 
        The A1 benchmark contains hourly traffic measurements from a web service; the figure marks manually labeled anomalies with red circles.
        From the 67 datasets in A1, we selected six time series in which a hold-out split—training $49\%$, validation $21\%$, and test $30\%$—contains at least one anomaly in every subset.
        Each panel plots training data in orange, validation data in blue, and test data in green.
        Task numbers match the IDs assigned in the A1 benchmark.}
    \label{RealData}
\end{figure}

To further evaluate the SR-RC architecture, we validated the Yahoo! Webscope S5 dataset, which is widely used for time-series anomaly detection \cite{ren2019time,laptev2015yahoo,zamanzadeh2024deep,kim2023time,yoshihara2022simple,hou2025network}.
The dataset is organised into A1–A4 benchmarks, where A1 consists of hourly traffic measurements from production web services with manually annotated anomalies. 
As the SR-RC architecture uses supervised learning, we selected the six time series shown in Fig. ~\ref{RealData} from the A1 benchmark (67 time series), which contains anomalies in all three splits when the data are partitioned using a hold-out scheme into training, validation, and test subsets ($49\%$, $21\%$, and $30\%$, respectively).
When treating these series as inputs, we applied min–max normalisation to match the scale of the saliency map.
All models—including SR-Logi and Multi-SR-Logi—are trained with class-weighted logistic loss; for the two proposed models and for RC, only the output coupling weights are optimised via class-weighted logistic regression.
Class weights are set inversely proportional to class frequencies so that misclassification of the minority class causes the loss function to assume larger values (see the Methods section).

Table~\ref{RealDataResults} summarises the learning results for Yahoo! Webscope S5 for the two proposed models, the conventional logistic-regression models, and RC.
Multi-SR-RC achieved the highest performance across the tasks. 
For some tasks, the performance aligned with that of RC and conventional logistic regression. 
We also observed that SR-RC can underperform RC and Multi-SR-Logi depending on the task.

\begin{table}[t]
\caption{\textbf{Performance evaluation on real-world time-series anomaly-detection tasks.}
Mean F1 score (mean $\pm$ s.d.), averaged over 10 runs. Columns 2–4 list conventional models, and columns 5–6 list proposed models. For each task, the best value is shown in \textbf{bold}; when all methods tie, no emphasis is applied.}
\centering
\begin{tabular}{cccccc}
\toprule
& \multicolumn{3}{c}{conventional models} & \multicolumn{2}{c}{proposed models} \\
\cmidrule(lr){2-4}\cmidrule(lr){5-6}
Task No. & SR-Logi & Multi-SR-Logi & RC & SR-RC & Multi-SR-RC \\
\midrule
13 & 0.580 $\pm$ 0.000 & 0.899 $\pm$ 0.000 & 0.954 $\pm$ 0.000 & 0.729 $\pm$ 0.035 & \textbf{1.000 $\pm$ 0.000}  \\
15 & 0.576 $\pm$ 0.000 & \textbf{1.000 $\pm$ 0.000} & \textbf{1.000 $\pm$ 0.000} & 0.741 $\pm$ 0.039 & \textbf{1.000 $\pm$ 0.000} \\
23 & 0.584 $\pm$ 0.000 & 0.862 $\pm$ 0.000 & \textbf{0.899 $\pm$ 0.000} & 0.588 $\pm$ 0.002 & \textbf{0.899 $\pm$ 0.000}  \\
29 & 0.832 $\pm$ 0.000 & 0.832 $\pm$ 0.000 & 0.832 $\pm$ 0.000 & 0.832 $\pm$ 0.000 & 0.832 $\pm$ 0.000  \\
52 & 0.719 $\pm$ 0.000 & 0.719 $\pm$ 0.000 & 0.719 $\pm$ 0.000 & 0.719 $\pm$ 0.000 & 0.719 $\pm$ 0.000  \\
60 & 0.586 $\pm$ 0.000 & 0.727 $\pm$ 0.000 & 0.876 $\pm$ 0.012 & 0.767 $\pm$ 0.016 & \textbf{0.929 $\pm$ 0.007} \\
\bottomrule
\end{tabular}
\label{RealDataResults}
\end{table}

\section*{Discussion}
We propose the SR-RC architecture, which integrates a saliency-map–driven bottom-up attention mechanism with RC and evaluates it on time-series anomaly detection benchmark tasks and the Yahoo! Webscope S5 dataset.
Specifically, the SR-RC architecture consists of two variants: SR-RC, which uses the saliency map as input to the RC, and Multi-SR-RC, which uses both the saliency map and original time-series data as inputs to RC.
The SR-RC architecture outperformed conventional logistic-regression models and a conventional RC model. Within the SR-RC architecture, Multi-SR-RC achieved the best performance, followed by SR-RC.

We first consider why introducing RC into the SR method improves performance.
Owing to its time-history effect, the RC can capture the characteristic features of time-series patterns, namely, the dependencies between a given time point and its past context \cite{lukovsevivcius2009reservoir,cucchi2022hands}.
Leveraging this effect, SR-RC can reach back prior to each time step’s value provided by the saliency map input, integrate the sequence information, and perform anomaly detection.
By contrast, because SR-Logi performs threshold-based detection using the magnitude of the saliency map values at each time step, it does not exhibit a time-history effect.
Accordingly, as shown in Fig.~\ref{internalState}, SR-RC can detect even subtle spikes that SR-Logi does not recognize by using information immediately before the spike, which likely explains SR-RC’s better performance.
Moreover, in SR-Logi, the classifier uses a single feature at each timestep.
In contrast, SR-RC can leverage nonlinear, high-dimensional spatiotemporal responses from all neurones, connecting the reservoir layer to the output layer, which also contributed to the performance improvement.

Second, we examine how adding the original time-series data to the saliency extracted by the SR method contributes to performance gains.
As shown in Table~\ref{RealDataResults}, SR-RC and SR-Logi, which use only the saliency map, perform worse than Multi-SR-Logi, conventional RC, and Multi-SR-RC. 
These latter three models use, respectively, the saliency map plus the original time series, the original time series alone, and both inputs.
We infer that this difference arises because of the limited information derived from the SR method.
The SR method primarily extracts anomalies that appear as salient deviations or change points, whereas anomalies distributed continuously over time (e.g. pattern-wise outliers) are difficult to extract adequately.
Therefore, combining the original time-series data can complement the anomaly information that is difficult to extract sufficiently using the SR method, which possibly led to performance gains.
Furthermore, through this complementation, Multi-SR-RC likely achieved the highest performance across all tasks by jointly integrating and evaluating sequence information from both the original time-series data and the saliency map.
In neural mechanisms, saliency information derived from sensory stimuli is integrated with the sensory stimuli to support efficient information processing \cite{chica2013two,bisley2019neural,moore2003selective,zhou2011feature}.
Accordingly, the mechanism of Multi-SR-RC has been explained using insights from neuroscience.

Third, we consider the applicability of SR-RC to Edge AI and its hardware implementation.
RC primarily requires two capabilities: mapping input signals to a high-dimensional state space and exploiting the time-history effect.
A time-delay dynamic system with delayed feedback to a single nonlinear element can realise these properties and its simple configuration facilitates hardware implementation \cite{tanaka2019recent}.
In such single nonlinear element implementations, the fine-grained tuning of an RNN’s internal connections is not feasible.
Consequently, models that require the learning of internal coupling weights, such as long short-term memory (LSTM) \cite{hochreiter1997long} and gated recurrent units (GRUs) \cite{cho2014learning}, are difficult to implement; however, RC, which fixes the internal recurrent weights, is well suited.
In particular, in actual implementations, spintronic oscillators and silicon photonic integrated circuits are effective candidates, as RC has already been demonstrated on such hardware \cite{tanaka2019recent,nakajima2020physical,marrows2024neuromorphic,van2017advances}.
In the SR method, the dominant computations are FFT and IFFT, which have been reported to achieve high energy efficiency when implemented on DSP or ASIC hardware \cite{garrido2022survey,garrido2020evolution,noor2018design}.
This processing remains on the device and avoids the transfer of data to remote servers.
By contrast, implementing top-down attention mechanisms, such as self-attention on DSPs or ASICs, can strain physical resources owing to large-scale matrix operations and memory demands, creating potential bottlenecks \cite{kang2024survey, leroux2025analog, bettayeb2024efficient}.
These considerations suggest that a bottom-up attention mechanism based on RC and the SR method aligns well with hardware implementation and motivates the deployment of SR-RC for edge AI relative to models such as LSTM, GRUs, and self-attention.

Finally, this study had some limitations.
Real-world anomaly detection tasks suggest the presence of anomalies that the SR method does not fully extract, indicating the need to improve SR-based extraction.
For example, although we currently compute the saliency map using FFT, replacing it with a wavelet transform—thereby enabling data-adaptive frequency resolution and temporal resolution—may allow for more detailed anomaly extraction.
It is necessary to examine this substitution in future research.
In addition, because SR-RC trains only the output layer, it remains unclear whether SR-RC can match other state-of-the-art approaches with the same energy consumption and physical hardware resources.
Consequently, future research should evaluate the detection performance, energy efficiency, FLOPs, and related aspects from multiple perspectives in actual hardware environments.

\section*{Conclusion}
In this study, we proposed an SR-RC architecture that integrates a bottom-up attention mechanism based on the SR method with RC and evaluated its performance on benchmark tasks and real-world datasets for time-series anomaly detection. 
Across these evaluations, the proposed models outperformed the conventional RC and logistic-regression model based on the values extracted using the SR method.
Accordingly, although there are several limitations, the SR-RC architecture is a promising approach for deploying RC as Edge AI for time-series anomaly detection.

\section*{Methods}

\subsection*{Spectral residual (SR) method}\label{SR}
The left panel of Fig.~\ref{ConceptualFigure_ProposedModel} presents an overview of the SR method.
We applied the SR method as a bottom-up attention mechanism to extract anomalies from the time series \cite{hou2007saliency}. 
Given $ \mathbf{U} = \left( U_{0}, U_{1}, \dots, U_{T-1} \right) \in \mathbb{R}^{T} $, the series is partitioned into overlapping windows of size $\tau$ with an overlap ratio $r$ ($0 \leq r < 1$). 
The step size is defined as $ s = \lfloor \tau (1-r) \rfloor$, and the total number of windows is $K = \left\lceil \frac{T-\tau}{s} \right\rceil + 1$, where \(\lceil \cdot \rceil\) denotes the ceiling function and $\lfloor \cdot \rfloor$ denotes the floor function.
In this study, $\tau = 128$ and $r = 0.5$.
For each segmented window, the \(k\)th window \(\mathbf{u}^{(k)}\) (where \(k = 1,\dots,K\)) is defined as 
\begin{align}
    \mathbf{u}^{(k)} =
    \begin{cases}
        \left( U_{t_k},\, U_{t_k+1},\, \dots,\, U_{t_k+\tau-1} \right) & \text{if } t_k+\tau-1 \leq T-1, \\
        \left( U_{t_k},\, U_{t_k+1},\, \dots,\, U_{T-1} \right) & \text{if } t_k+\tau-1 > T-1,
    \end{cases}
\end{align}
with the starting position $t_k = s(k-1)$. 
If the window lies entirely within \(\mathbf{U}\), then \(\mathbf{u}^{(k)} \in \mathbb{R}^{\tau}\). 
Otherwise, only data up to \(U_{T-1}\) are used, and the excess steps are \(\Delta_k = (t_k + \tau) - T\), yielding \(\mathbf{u}^{(k)} \in \mathbb{R}^{\tau - \Delta_k}\). 
We define \(L_k\) as the length of the \(k\)-th window; thus, \(L_k = \tau\) when \(t_k + \tau - 1 \leq T - 1\) and \(L_k = \tau - \Delta_k\) otherwise.
Let $u^{(k)}_t=U_{t_k+t}$ be the $t$th component of the $k$th window, for $t=0,1,\dots,L_k-1$.
The discrete Fourier transform and its inverse are defined as 
\begin{align}
    X^{(k)}_{f} &= \mathfrak{F}\{ u^{(k)}_t \} = \sum_{t=0}^{L_k-1} u^{(k)}_t\, e^{-i \frac{2\pi f t}{L_k}}, \quad f = 0, 1, \dots, L_k-1, \\
    u^{(k)}_t &= \mathfrak{F}^{-1}\{ X^{(k)}_{f} \} = \frac{1}{L_k}\sum_{f=0}^{L_k-1} X^{(k)}_{f}\, e^{i \frac{2\pi f t}{L_k}}, \quad t = 0, 1, \dots, L_k-1.
\end{align}
Here, $\mathfrak{F}$ denotes the discrete Fourier transform, and $\mathfrak{F}^{-1}$ represents the inverse discrete Fourier transform.
The amplitude and phase spectra were obtained as follows:
\begin{align}
    A_f = |X^{(k)}_{f}|, \quad P_f = \arg(X^{(k)}_{f}).
\end{align}
Anomalies in the time series data, such as outliers or trend changes, appear as abrupt fluctuations and are reflected as noise-like variations distributed across the entire amplitude spectrum $A_f$.
Because these variations indicate anomalies, the SR method identifies them by subtracting the moving average of the log-amplitude spectrum, which is computed over a window of size $q$ (where $q=3$) from the log-amplitude spectrum.
This difference is defined by the spectral residual \(R_f \in \mathbb{R}\):
\begin{align}
    &\{R_0,\dots,R_f,\dots,R_{L_k-1}\} \\
    &= \log(\{A_0,\dots,A_f,\dots,A_{L_k-1}\}) - \mathbf{h}_q * \log(\{A_0,\dots,A_f,\dots,A_{L_k-1}\}),\notag\\ 
    &\mathbf{h}_q = \frac{1}{q}\begin{bmatrix}1 & 1 & \dotsb & 1\end{bmatrix},
\end{align}
where \(*\) denotes the convolution operation and \(\mathbf{h}_q\in \mathbb{R}^q\) is the convolution kernel used to compute the local average of $\log(A_f)$.
Finally, by combining \(R_f\) with the phase spectrum \(P_f\) (which retains the original phase information) and applying an inverse Fourier transform, we obtain the saliency map $S(\mathbf{u}^{(k)}) \in \mathbb{R}^{L_k}$.
\begin{align}
 &S(\mathbf{u}^{(k)})
 = \left| \mathfrak{F}^{-1} \bigl( \exp (R_f+iP_f) \bigr) \right|.
\end{align}
When using these saliency maps as input signals to RC, we concatenate $S(\mathbf{u}^{(1)}),\dots,S(\mathbf{u}^{(K)})$ horizontally and average values over overlapping regions, yielding a saliency map $\mathbf{S} \in \mathbb{R}^{T}$ that matches the dimensions of the original time-series data.
For each time index $j$ $(j=0,1,\dots,T-1)$, define the set of windows that cover $j$ by $\mathcal{K}(j)=\{\,k\in\{1,\ldots,K\}\mid 0\le j-t_k<L_k\,\}$.
We then define $S_j\in\mathbb{R}$ as the average value of the saliency map at time $j$ across all windows that cover $j$ (if a single window covers $j$, we use that value directly):
\begin{align}
    S_j=\frac{1}{|\mathcal{K}(j)|}\sum_{k\in \mathcal{K}(j)} S\bigl(\mathbf{u}^{(k)}|_{j-t_k}\bigr).
\end{align}
Finally, by concatenating the values at each time step, we constructed the saliency map $\mathbf{S}$ as
\begin{align}
    \mathbf{S}=[S_0,S_1,\ldots,S_j,\ldots,S_{T-1}].
\end{align}

\subsection*{Model}

\subsubsection*{Proposed models}
The frameworks of SR-RC and Multi-SR-RC are shown in the bottom-right panel of Fig. ~\ref{ConceptualFigure_ProposedModel}.
The definitions of the internal states are expressed in Eqs.~(\ref{eq:SR-RC}) and (\ref{eq:Multi-SR-RC}), respectively.
The outputs of the SR-RC and Multi-SR-RC at time $t$ are obtained using logistic regression.
Specifically, using the internal states from each reservoir layer and output coupling weights, the output is defined as
\begin{align}
    y_{t}=\frac{1}{1+\exp\{-(\mathbf{W}_{\mathrm{out}}\mathbf{x}_{t}+b)\}} . \label{eq.out12}
\end{align}
Here, $y_t\in\mathbb{R}$ is a one-dimensional output signal, $\mathbf{W}_{\mathrm{out}}\in\mathbb{R}^{N}$ denotes the output coupling weights, and $b\in\mathbb{R}$ is a bias term.
In this setting, $\mathbf{W}_{\mathrm{out}}$ is optimised using the maximum likelihood method (described below).
Finally, we define the final output $\hat{y}_t \in \{0,1\}$ by thresholding $y_t$ at $\Theta$ (here, $\Theta=0.5$):
\begin{align}
    \hat{y}_t =
        \begin{cases}
            1, & \text{if } y_t \ge \Theta, \\
            0, & \text{if } y_t < \Theta.
        \end{cases}
        \label{eq.13}
\end{align}

\subsubsection*{Reservoir computing (RC) model} \label{RC}
In the conventional RC model (upper-right panel of Fig.\ref{ConceptualFigure_ProposedModel}), the network comprises input, reservoir, and output layers. 
The temporal evolution of a conventional RC can be obtained from a Multi-SR-RC by removing the saliency map input. 
At time $t$ ($0\le t\le T$), the reservoir dynamics are defined as
\begin{align} 
    \mathbf{x}_t = (1-\alpha)\,\mathbf{x}_{t-1} + \alpha \cdot \tanh\left(\mathbf{W}_{\mathrm{in}}\,u_t + \mathbf{W}\,\mathbf{x}_{t-1}\right).
\end{align} 
The output at time \(t\) is given as
\begin{align} 
   y_{t} = \frac{1}{1+\exp\{-({\mathbf{W}_{\mathrm{out}}\,\mathbf{x}_{t}}+b)\}},\label{eq.out14}
\end{align} 
As in SR-RC and Multi-SR-RC, $\mathbf{W}_{\mathrm{out}}$ is optimised using maximum-likelihood estimation.
Finally, as in Eq.~(\ref{eq.13}), we obtain the final output $\hat{y}_t \in \{0,1\}$ by thresholding $y_t$ at $\Theta$ (here, $\Theta=0.5$).

\subsubsection*{Parameter settings} \label{sec:setParam}
We configured the recurrent coupling weights $\mathbf{W}$, input coupling weights for the original time-series data $\mathbf{W}_{\mathrm{in}}$, and input coupling weights for the saliency map $\mathbf{W}_{\mathrm{S}}$ as follows:
$\mathbf{W}$ is determined by the sparsity rate $\beta (0 < \beta \le 1)$ and spectral radius $\gamma (0 < \gamma)$.
Prior to adjusting the recurrent coupling weights, a uniform matrix $\mathbf{W}_{0} \in  \mathbb{R}^{N \times N}$ with a mean of zero is prepared.
The matrix is then sparsified by setting the elements to zero at a sparsity rate $\beta$.
Finally, the recurrent coupling weight matrix $\mathbf{W}$ is obtained by scaling the sparsified matrix:
    \begin{align} 
    \mathbf{W} = \gamma\frac{\mathbf{W}_0}{\rho(\mathbf{W}_0)},
    \end{align} 
where $\rho(\mathbf{W}_0)$ denotes the largest eigenvalue of $\mathbf{W}_{0}$.
In addition, each element of $\mathbf{W}_{\mathrm{in}}$ is drawn uniformly from the interval $[-a^\mathrm{in}, a^\mathrm{in}]$, where $a^\mathrm{in}$ is the input scaling parameter.
Similarly, each element of $\mathbf{W}_{\mathrm{S}}$ is drawn uniformly from $[-a^{\mathrm{S}}, a^{\mathrm{S}}]$, where $a^{\mathrm{S}}$ is the scaling parameter.

\subsubsection*{Classifier for comparison: logistic regression based on saliency maps and original time-series data}
In this study, we compared the proposed models with conventional approaches for anomaly detection based on logistic regression, namely SR-Logi and Multi-SR-Logi.
This evaluation was conducted to verify the suitability of RC as a classifier.

In SR-Logi, at time $t\,(0 \le t \le T)$, the probability of an anomaly occurrence, \(y_t \in [0,1]\), corresponding to the one-dimensional saliency map \(S_t \in \mathbb{R}\) is defined as follows:
\begin{align}
    y_t = \frac{1}{1+\exp\{-(\beta_0+\beta_1 S_t)\}},\quad t=0,1,\dots,T,
\end{align}
where \(\beta_0 \in \mathbb{R}\) denotes the intercept and \(\beta_1 \in \mathbb{R}\) is the regression coefficient for \(S_t\).
These parameters were estimated using maximum likelihood estimation. 
Based on \(y_t\), the final prediction of whether an anomaly occurs, \(\hat{y}_t \in \{0,1\}\), is determined according to the threshold \(\Theta\) (\(0 \le \Theta \le 1\)) as follows:
\begin{align}
    \hat{y}_t =
        \begin{cases}
            1, & \text{if } y_t \ge \Theta, \\
            0, & \text{if } y_t < \Theta,
        \end{cases}
        \quad t=0,1,\dots,T,
\end{align}
where \(\Theta\) is optimised using a Bayesian optimization \cite{snoek2012practical}.

In Multi-SR-Logi, at time \(t\), both the saliency map \(S_t\) and original time-series data \(u_t \in \mathbb{R}\) are used as features. 
The probability of an anomaly occurrence, \(z_t \in [0,1]\), is defined as
\begin{align}
    z_t = \frac{1}{1+\exp\{-(\beta_0+\beta_1 S_t+\beta_2 u_t)\}},\quad t=0,1,\dots,T,
\end{align}
where \(\beta_2 \in \mathbb{R}\) is the regression coefficient for \(u_t\), which is optimised using the maximum likelihood estimation in the same manner as in SR-Logi. 
Based on \(z_t\), the final prediction of whether an anomaly occurs, \(\hat{z}_t \in \{0,1\}\), is defined by the following equation:
\vspace{-1ex}
\begin{align}
    \hat{z}_t =
    \begin{cases}
        1, & \text{if } z_t \ge \Theta, \\
        0, & \text{if } z_t < \Theta,
    \end{cases}
    \quad t=0,1,\dots,T.
\end{align}

\subsection*{Learning procedure}
The output weights \(\mathbf{W}_{\mathrm{out}}\) of SR-RC, Multi-SR-RC, and the conventional RC model were optimized by maximum-likelihood estimation.
Using the model output $y_t$ at time $t$ as defined in Eqs.~(\ref{eq.out12}) and (\ref{eq.out14}) for the respective models, together with the teacher signal $d_t \in \{0,1\}$, we define the loss function parameterized by $\mathbf{W}_{\mathrm{out}}$ as 
\begin{align}
L(\mathbf{W}_{\mathrm{out}}) 
= - \sum_{t=0}^{T} \bigl[ d_t \ln y_t + (1 - d_t) \ln (1 - y_t) \bigr].
\end{align}
Minimising \(L(\mathbf{W}^{\mathrm{out}})\) yields the optimised weights \(\mathbf{W}_{\mathrm{out}}^\ast\) as follows: 
\begin{align}
\mathbf{W}_{\mathrm{out}}^\ast 
= \operatorname*{arg\,min}_{\mathbf{W}_{\mathrm{out}}} L(\mathbf{W}_{\mathrm{out}}).
\end{align}

In a real-world time-series anomaly detection task, we apply weighting to the loss function to handle a smaller number of anomalies.
The weights $w_1\in \mathbb{R}$ and $w_0 \in \mathbb{R}$ are inversely proportional to the occurrence frequencies of anomalous and normal data, respectively, and are defined as follows:
\begin{align}
&w_1=\frac{n_1+n_0}{2\,n_1},\qquad 
w_0=\frac{n_1+n_0}{2\,n_0},\qquad\\
&n_1=\sum_{t=0}^{T-1}\mathbbm{1}\{d_t=1\},\quad n_0=\sum_{t=0}^{T-1}\mathbbm{1}\{d_t=0\}.
\end{align}
Here, $\mathbbm{1}(\cdot)$ is the indicator function and equal to $1$ when its argument is true, otherwise $0$.
The weighted loss function $J(\boldsymbol{W}^{\mathrm{out}})$ is defined as follows:
\begin{align}
    J(\boldsymbol{W}^{\mathrm{out}})=-\sum_{t=0}^{T-1}\left[w_1 d_t \ln \hat{y}_t + w_0 (1-d_t)\ln(1-\hat{y}_t)\right].
\end{align}

\subsection*{Datasets of synthetic time-series anomaly-detection benchmark task}
Lai et al. introduced a benchmark that evaluates time-series anomaly detection by injecting anomalies into the baseline time-series data \cite{Lai2021RevisitingTS}. 
Anomalies were classified as either point-wise or pattern-wise. 
We adopted this benchmark and described its three components as baseline time-series data, point-wise outliers, and pattern-wise outliers.

\subsubsection*{Baseline time series data}
Let \(\mathbf{u} = (u_{1},u_{2},\dots,u_{T}) \in \mathbb{R}^{T}\) denote the baseline time series where each \(u_t \in \mathbb{R}\) (for \(t = 1, \dots, T\)) is defined as
\begin{align}    
    &u_t = \sum_{p=1}^{P}[\mathrm{A\,sin}(2\pi f_{p}t+\phi_p)+\mathrm{B\,cos}(2\pi f_{p}t+\phi_p)] + \epsilon(t),\,\, t=1,\dots,T,
\end{align}
where \(P\) is the number of superimposed waves; \(\mathrm{A}\) and \(\mathrm{B}\) are coefficients that determine the amplitude; \(f_p\) is the frequency of the $p$th wave; $\phi_p$ is the phase; and \(\epsilon(t)\sim\mathcal{N}(0, 0.05)\) is the noise term.
For the single-sinusoid case, we set \(A=1\), \(B=0\), \(P=1\), \(f_{1}=0.04\), and \(\phi_{1}=0\). 
For the four-sinusoid case, we set \(A=1\), \(B=0\), \(P=4\), \(f_{1}=0.005\), \(f_{2}=0.015\), \(f_{3}=0.02\), \(f_{4}=0.04\), and \(\phi_{1}=0\), \(\phi_{2}=\pi/8\), \(\phi_{3}=\pi/4\), \(\phi_{4}=\pi/2\). 
When using irrational frequency ratios, we set \(f_{1}=\sqrt{2}\), \(f_{2}=\sqrt{5}\), \(f_{3}=\sqrt{7}\), and \(f_{4}=\sqrt{11}\). 
We set \(T=3000\) for all tasks.

\subsubsection*{Point-wise outliers}
Point-wise outliers are defined as anomalies that appear as extreme values at specific time points or values that deviate significantly from the surrounding data points in a time series. 
In this study, two types of pointwise outliers are considered: global outliers (illustrated in the lower-right panel of Fig.\ref{ConceptualFigure_BenchMarkTask}) and contextual outliers (shown in the upper-right panel of Fig.\ref{ConceptualFigure_BenchMarkTask}).

The time steps at which these outliers replace the original values in the time-series data \(\mathbf{u}=(u_{1},u_{2},\ldots,u_{T})\) are determined using a Poisson process that assigns an anomaly occurrence probability \(\delta\) ($0<\delta \le 1$) to each time step. 
The set of time points at which anomalies occur is denoted by \(\mathbf{i} = (i_1,i_2,\dots,i_m,\dots,i_M)\), where \(i_m \in \{1,\dots,T\}\).

Global outliers are points that deviate markedly from the values at other time steps and typically appear as spikes in the time series.
They are generated by replacing the value \(u_{i_m}\) at each time step \(i_m\) \((m=1,\dots,M)\) with \(\hat{u}_{i_m} \in \mathbb{R}\) according to
\begin{align}
    \hat{u}_{i_m} = \mu(\mathbf{u}) \pm \lambda \cdot \sigma(\mathbf{u}), \quad m=1,2,\dots,M,
\end{align}
where \(\mu(\mathbf{u})\) is the mean of the time series \(\mathbf{u}\), \(\sigma(\mathbf{u})\) is the standard deviation, and \(\lambda\) is a parameter that controls the anomaly magnitude. 
In this study, we set \(\lambda =3.5\).

Contextual outliers are defined as anomalous values \(u'_{i_m}\) that deviate from the mean and standard deviations computed over a local neighborhood centered at time point \(i_m\) in \(\mathbf{u}\). 
This local neighborhood is defined as \(\mathbf{u}'_m=(u_{i_m-k},\ldots,u_{i_m+k})\). 
For each \(i_m\), \(u'_{i_m}\) is given by
\begin{align}
u'_{i_m} = \mu(\mathbf{u}'_m) \pm \lambda \cdot \sigma(\mathbf{u}'_m), \quad m=1,\dots,M,
\end{align}
where \(\mu(\mathbf{u}')\) and \(\sigma(\mathbf{u}')\) denote the mean and standard deviation of \(\mathbf{u}'\), respectively. 
In this study, we set \(k=5\) steps.

\subsubsection*{Pattern-wise outliers}
Pattern-wise outliers are defined as anomalies in which a consecutive segment of data exhibits characteristics or shapes that are distinctly different from the original pattern. 
In this study, attern-wise outliers were categorised into two types: seasonal outliers (shown in the upper-left panel of Fig.\ref{ConceptualFigure_BenchMarkTask}) and shapelet outliers (shown in the lower-left panel of Fig.\ref{ConceptualFigure_BenchMarkTask}).

The starting times for the segments to be replaced by pattern-wise outliers are generated from time-series data \(\mathbf{u}=(u_{1},u_{2},\dots,u_{T})\) according to a Poisson process, where each time step has an anomaly occurrence probability \(\delta\). 
The set of starting times for the pattern-wise outliers is denoted by \(\mathbf{j}=(j_1,j_2,\dots,j_{m},\dots,j_{M'})\) with \(j_{m}\in \{1,\dots,T\}\).

Shapelet outliers are produced by replacing the segment of the time series \(\mathbf{u}\) that begins at time step \(j_m\) and spans \(k'\) time steps with an alternative waveform pattern featuring a modified amplitude, denoted by \(\tilde{\mathbf{u}}_m=(\tilde{u}_{j_m},\tilde{u}_{j_m+1},\dots,\tilde{u}_{j_m+k'})\). 
At any time \(t'\) in \(\tilde{u}_{t'}\), where \(t'=j_m,j_m+1,\dots,j_m+k'\), the value \(\tilde{\mathbf{u}}_m\) is defined as

\begin{align}
    \tilde{u}_{t'} = \sum_{n'=0}^{N'-1}\Big[A' \sin(2\pi \tilde{f}_{n'}t') + B' \cos(2\pi \tilde{f}_{n'}t')\Big] + \epsilon'(t'),\quad t'=j_m,j_m+1,\dots,j_m+k' ,
\end{align}
where \(\epsilon'(t')\) denotes the noise term sampled from a normal distribution with mean \(0\) and variance \(1\). 
In this study, we set \(k=20\) time steps and superimpose \(N'=5\) waves and employ rectangular sine waves characterised by \(A' = A/(2n'+1)\), \(B'=0\), and \(\tilde{f}_{n'} = 0.04(2n'+1)\).

For seasonal outliers, anomalies are introduced by replacing the segment of the time series \(\mathbf{u}\) that begins at time step \(j_m\) and spans \(k'\) time steps with a new segment \(\check{\mathbf{u}}_m = (\check{u}_{j_m},\check{u}_{j_m+1},\dots,\check{u}_{j_m+k'})\), which is characterized by a different set of frequencies \(f'_{n}\). 
At any time \(t'\) in \(\check{u}_{t'}\), the value \(\check{u}_{t'}\) is defined as
\begin{align}
    \check{u}_{t'} &= \sum_{p=1}^{P}\Big[A \sin(2\pi f'_{p}t') + B \cos(2\pi f'_{p}t')\Big],\quad t'=j_m,j_m+1,\dots,j_m+k'\\
    f'_{p} &= \lambda' \cdot f_{p}, \notag
\end{align}
where \(\lambda'\) is the frequency-scaling parameter, which was set as \(\lambda' = 3.5\) in this study.

\subsection*{Datasets of real-world time-series anomaly-detection task}
For the real-world time-series anomaly-detection task, we used the Yahoo! Webscope S5 dataset \cite{laptev2015yahoo}.
This dataset is widely used for time-series anomaly detection \cite{ren2019time,zamanzadeh2024deep,kim2023time,yoshihara2022simple,hou2025network}.
We used the A1 benchmark of Yahoo! Webscope S5 (hourly web traffic) and selected six of its 67 series ( Fig. ~\ref{RealData}), which retained anomalies across the hold-out splits into training, validation, and test sets ($49/21/30\%$).
The dataset was publicly available at the time of use and is not currently available.

\subsection*{Evaluation method}
In this study, the mean F1 score was used as the evaluation metric for the proposed model. 
This metric enables the assessment of the model’s performance in distinguishing between normal and anomalous conditions.
True positive (TP) is defined as the number of anomalous instances correctly identified as anomalous, false positive (FP) as the number of normal instances incorrectly classified as anomalous, false negative (FN) as the number of anomalous instances incorrectly classified as normal, and true negative (TN) as the number of normal instances correctly identified as normal.
Next, we calculated the F1 scores for the normal and anomalous classes, denoted by $F1_{\mathrm{norm}}$ and $F1_{\mathrm{an}}$, respectively.
These are defined as follows:
\begin{align}
F1_{\mathrm{norm}} = \frac{2TN}{2TN + FP + FN},\\
F1_{\mathrm{an}} = \frac{2TP}{2TP + FP + FN}.
\end{align}
The mean F1 score is defined as the average of the two scores:
\begin{align}
\text{Mean F1 score} = \frac{F1_{\mathrm{norm}} + F1_{\mathrm{an}}}{2}.
\end{align}
A higher mean F1 score indicates better performance.

\subsection*{Experimental environment}
Table \ref{tab:params} summarises the parameters used in all experiments, including those for the proposed models—Multi-SR-RC and SR-RC—as well as for the conventional logistic-regression models (SR-Logi and Multi-SR-Logi), the conventional RC model, and the benchmark tasks.
At this stage, several parameters in the proposed models, conventional RC model, and logistic-regression models are optimised using Bayesian optimisation \cite{frazier2018tutorial,snoek2012practical}. 
Specifically, we tuned the spectral radius $\gamma$, leak rate $\alpha$, input-scaling parameter for the original time-series data $a^\mathrm{in}$, input-scaling parameter for the saliency map $a^\mathrm{s}$, and threshold $\Theta$ in the logistic-regression models.

\begin{table}[t]
\caption{\textbf{Parameter settings for all experiments}}
\label{tab:params}
\centering
\begin{tabular*}{\textwidth}{@{\extracolsep{\fill}}l l l}
\toprule
Model & Hyperparameter & Setting / search space \\
\midrule
\multicolumn{3}{@{}l}{\textbf{Proposed models}} \\
\addlinespace[2pt]
SR-RC &
Input weight scale for the saliency map $a^\mathrm{s}$ & $0.01-5$\footnotemark[1] \\
& Spectral radius $\gamma$ & $0.01-3$\footnotemark[1] \\
& Leak rate $\alpha$ & $0-1$\footnotemark[1] \\
& Sparsity $\beta$ & $0.01-1$\footnotemark[1] \\
& Reservoir size $N$ & $100$ \\
& Threshold $\Theta$ & $0.5$\\
\addlinespace
Multi-SR-RC &
Input weight scale for original series $a^\mathrm{in}$ & $0.01-5$\footnotemark[1] \\
& Other hyperparameters & same as SR-RC \\
\midrule
\multicolumn{3}{@{}l}{\textbf{Conventional models}} \\
\addlinespace[2pt]
RC & Same as Multi-SR-RC, excluding $a^\mathrm{s}$  \\
SR-Logi / Multi-SR-Logi & Threshold $\Theta$ & $0.01-1$\footnotemark[1] \\
\midrule
\multicolumn{3}{@{}l}{\textbf{Parameter for benchmark task}} \\
\addlinespace[2pt]
— & Anomaly occurrence probability $\delta$ & $0.05-0.3$ \\
\botrule
\end{tabular*}

\footnotetext[1]{Optimized via Bayesian optimization within the indicated range \cite{frazier2018tutorial,snoek2012practical}.}
\end{table}

\section*{Data availability}
The datasets generated and analysed during the current study, as well as the computer codes, are available from the corresponding author upon reasonable request.

\bibliography{paper}

\section*{Acknowledgements}
This work was partially supported by JSPS KAKENHI Grant-in-Aid for Transformative Research Areas (A) Grant Number JP25H02626 (SN), Grant-in-Aid for Scientific Research(B) Grant Number JP25K03198 (SN), JST PRESTO Grant Number JPMJPR22C5,
Project JPNP14004, commissioned by the New Energy and Industrial Technology Development Organization (NEDO),
JST Moonshot R\&D Grant Number JPMJMS2021, 
Institute of AI and Beyond of UTokyo, 
the International Research Center for Neurointelligence (WPI-IRCN) at The University of Tokyo Institutes for Advanced Study (UTIAS), 
and Cross-ministerial Strategic Innovation Promotion Program (SIP), the 3rd period of SIP ``Smart energy management system'' Grant Number JPJ012207.

\section*{Author contributions}
H.N., S.N., Y.S., and K.A. designed the study and analysed the results. 
H.N. and S.N. wrote the main manuscript text. 
H.N. prepared the figures and conducted the experiments. 
All authors reviewed and revised the manuscript.

\section*{Competing interests}
The authors declare no competing interests.

\end{document}